\documentclass[5pt]{article}
\usepackage{epstopdf}
\usepackage{graphicx}
\usepackage{tabularx}
\usepackage{multirow}
\usepackage[letterpaper]{geometry}
\usepackage{spconf,amsmath,epsfig}
\usepackage[ruled]{algorithm2e}
\usepackage{enumitem}
\usepackage{color}
\usepackage{fancyhdr}
\begin{document}\sloppy
\def\x{{\mathbf x}}
\def\L{{\cal L}}

\title{Adaptive Co-weighting Deep Convolutional Features \\ for Object Retrieval}
\name{Jiaxing Wang$^1$, Jihua Zhu$^1$, Shanmin Pang*$^1$, Zhongyu Li$^2$, Yaochen Li$^1$ and Xueming Qian$^3$ \thanks{*Corresponding author. }}
\address{${}^1$School of Software Engineering, Xi'an Jiaotong University, Xi'an, China\\${}^2$ Department of Computer Science, University of North Carolina at Charlotte, USA\\${}^3$School of Electronic and Information Engineering, Xi'an Jiaotong University, Xi'an, China\\{csuwjx}@stu.xjtu.edu.cn, {\{zhujh,pangsm\}@xjtu.edu.cn}\\zhongyu.emerald@gmail.com, {\{yaochenli,qianxm\}@xjtu.edu.cn}}
\maketitle

\begin{abstract}
Aggregating deep convolutional features into a global image vector has attracted sustained attention in image retrieval. In this paper, we propose an efficient unsupervised aggregation method that uses an adaptive Gaussian filter and an element-value sensitive vector to co-weight deep features. Specifically, the Gaussian filter assigns large weights to features of region-of-interests (RoI) by adaptively determining the RoI's center, while the element-value sensitive channel vector suppresses burstiness phenomenon by assigning small weights to feature maps with large sum values of all locations. Experimental results on benchmark datasets validate the proposed two weighting schemes both effectively improve the discrimination power of image vectors. Furthermore, with the same experimental setting, our method outperforms other very recent aggregation approaches by a considerable margin.
\end{abstract}

\section{Introduction}
\label{sec:intro}
When given a query image of an object, we are interested in finding images containing the same object from a large-scale database based on Content-based image retrieval (CBIR). The key step of CBIR is to generate an image representation, which involves aggregating patch-level feature descriptors into a single fixed-length image-level vector. Most state-of-the-art representations are based on hand-crafted features (e.g.~SIFT~\cite{lowe2004}) or Convolutional Neural Network ~\cite{krizhevsky2012} features.

The pioneer image representation model based on hand-crafted features is bag-of-word (BoW)~\cite{sivic2003video}, which maps each feature into a visual word and consequently represents an image as a high-dimensional sparse vector. BoW makes a celebrated success in CBIR, and numerous works~\cite{philbin2007object,  philbin2008lost, jegou2010improving, pang2014exploiting,arandjelovic2012three} adopt this model for the purpose of searching similar images. Although BoW has attracted great attention, it suffers two major drawbacks on large-scale retrieval: search efficiency and memory cost~\cite{jegou2010aggregating, li2017large}. An alternative solution is to aggregate local descriptors into a mid-size vector, e.g. Fisher vectors~\cite{perronnin2010large}, Vector of locally aggregated descriptors~\cite{jegou2010aggregating}, and etc~\cite{murray2017interferences, Pang2017beyond}.

Recently, the focus of image retrieval has shifted from hand-crafted features to CNN-based ones since the discriminative power of the latter are much stronger. Early works~\cite{babenko2014neural, gong2014multi, sharif2014cnn} consider the outputs of last fully-connected layer as global representations. After that, most recent works with superior performances, such as SPoC~\cite{babenko2015aggregating}, R-MAC~\cite{tolias2016particular} and CroW~\cite{kalantidis2016cross}, overall first employ the outputs of deep convolutional layer as local features, and then aggregate them into the global representation. Specifically, SPoC leverages centering prior to aggregate features output from the last convolutional layer with sum-pooling. It's worth noting the assumed centering prior that RoI is located at the geometric center of an image is probably not true for many images. Simply speaking, R-MAC first derives representations for image regions by performing max-pooling on the convolutional layer activations over the corresponding regions, and then calculates the image representation by aggregating region vectors with sum-pooling. Similar to SPoC, CroW also favors sum-pooling to directly aggregate convolutional layer features. The major difference between them is CroW employs a more effective cross-dimensional weighting strategy to weight deep features.
In more detail, besides performing spatial-wise weighting on feature maps, CroW also employs a channel-wise weighting to jointly boost discriminative power of features.

In this paper, we present an unsupervised deep feature weighting method to improve image representations. Our method is somewhat similar to CroW as we both perform spatial- and channel-wise weighting for each feature, but there are at least two distinct differences between them. First, our spatial weighting strategy, which extends SPoC by adaptively determining the center point of RoI, assigns larger weights to RoI features with an adaptive Gaussian (aGaussian) filter, while CroW leverages aggregated spatial response map to compute spatial-wise weight for each feature. Second, the proposed element-value sensitive channel (eChannel) weighting strategy obtains channel weights based on aggregating the product values of feature maps and aGaussian, while CroW derives channel weights with the sparsity of feature maps.
To summarize, our contributions are two-fold:
\begin{itemize}
	\item We design a strategy to adaptively determine the center point of RoI, and then incorporates this prior into Gaussian filter for assigning larger weights to RoI features.
	\item We design an eChannel weighting vector by aggregating the product values of feature maps and aGaussian for easing the intra-image visual burstiness~\cite{jegou2009burstiness}.
\end{itemize}
The organization of this paper is as follows. We review recent advances of CBIR, and outline our contributions in this section. Section 2 describes our two weighting strategies in detail. We support our method by extensive experiments in Section~\ref{Experiments}. Finally, we conclude our paper briefly in Section~\ref{Conclusion}.

\begin{algorithm}[t]
\caption{Deep feature aggregation framework}
\newcommand\mycommfont[1]{\textcolor{blue}{#1}}
\SetCommentSty{mycommfont}
\textbf{Input:} Tensor $X$, dimensionality $K^{'}$, parameter $\alpha$,whitening ~parameters $W$, aGaussian generation function ${f_G}$,eChannel generation function ${f_K}$ \\
\textbf{Output:} $K^{'}$-D global representation $\beta  \in {\mathcal{R}^{{K^{'}}}}$

	$S = {f_G}(X,\alpha)$ \tcp*{Adaptive Gaussian filter}

	${\Omega _k} = \sum\limits_{i = 1}^H {\sum\limits_{j = 1}^W {{X_{(k,i,j)}}} } {S_{(i,j)}}$ \tcp*{Spatial weighting}

	$B = {f_K}(X,S)$  \tcp*{Element-value vector}

	${\beta ^{'}} = B \odot \Omega$ \tcp*{Channel weighting}

	$\beta  = norm({\beta ^{'}})$ \tcp*{Normalize}

	${\beta ^{'}} = PCA({\beta ^{'}},W,{K^{'}})$ \tcp*{Whitening} 

	$\beta  = norm({\beta ^{'}})$       \tcp*{Normalize again} 
\end{algorithm}

\section{Methodology}
\subsection{Framework}
Let $X \in {\mathcal{R}^{( W \times H \times K)}}$ be the feature tensor extracted from the deep convolutional layer, which consists of $K$ feature maps (each having  width $W$ and height $H$). Let $S \in {\mathcal{R}^{(W \times H)}}$ and $B \in {\mathcal{R}^{K}}$ denote the spatial weighting matrix and the channel weight vector, respectively. We summarize the aggregation framework in Algorithm~1, where $\odot$ denotes element-wise product between vectors. In the following, we discuss our strategies of obtaining $S$ and $B$ in detail.

\subsection{Adaptive Gaussian filter for spatial weighting}
For effective object retrieval, it is better to pay more attention to RoI. In other words, The aggregation process should distinguish features of RoI and cluttered regions by assigning different weights to them:
\begin{eqnarray}{\Omega _k}{\rm{ = }}\sum\limits_{i = 1}^H {\sum\limits_{j = 1}^{\rm{W}} {{X_{(i,j,k)}}{S_{(i,j)}}} } {\kern 1pt} {\kern 1pt} ~\forall k=1,2,...,K
\end{eqnarray}
where ${S_{(i,j)}}$ denotes the weighting function. This function should assign large weights to RoI, and small weights to other regions. Accordingly, the probability density of Gaussian distribution can be selected for the purpose:
\begin{eqnarray}
{S_{(i,j)}} = \frac{1}{{2\pi {\sigma ^2}}}\exp \{  - \frac{{{{(i - {i_0})}^2} + {{(j - {j_0})}^2}}}{{2{\sigma ^2}}}\}
\end{eqnarray}
where $\sigma$ and $({i_{0}},{j_{0}})$ denote the standard deviation and center (mean value) of Gaussian distribution, respectively. For $\sigma$, we can set it to be the half distance between the geometrical center of the feature map and the farthest boundary.The problem arising here is how to determine $({i_{0}},{j_{0}})$.

\begin{figure}[t]	
	\begin{minipage}[b]{1.0\linewidth}
		\centering
		\centerline{\epsfig{figure=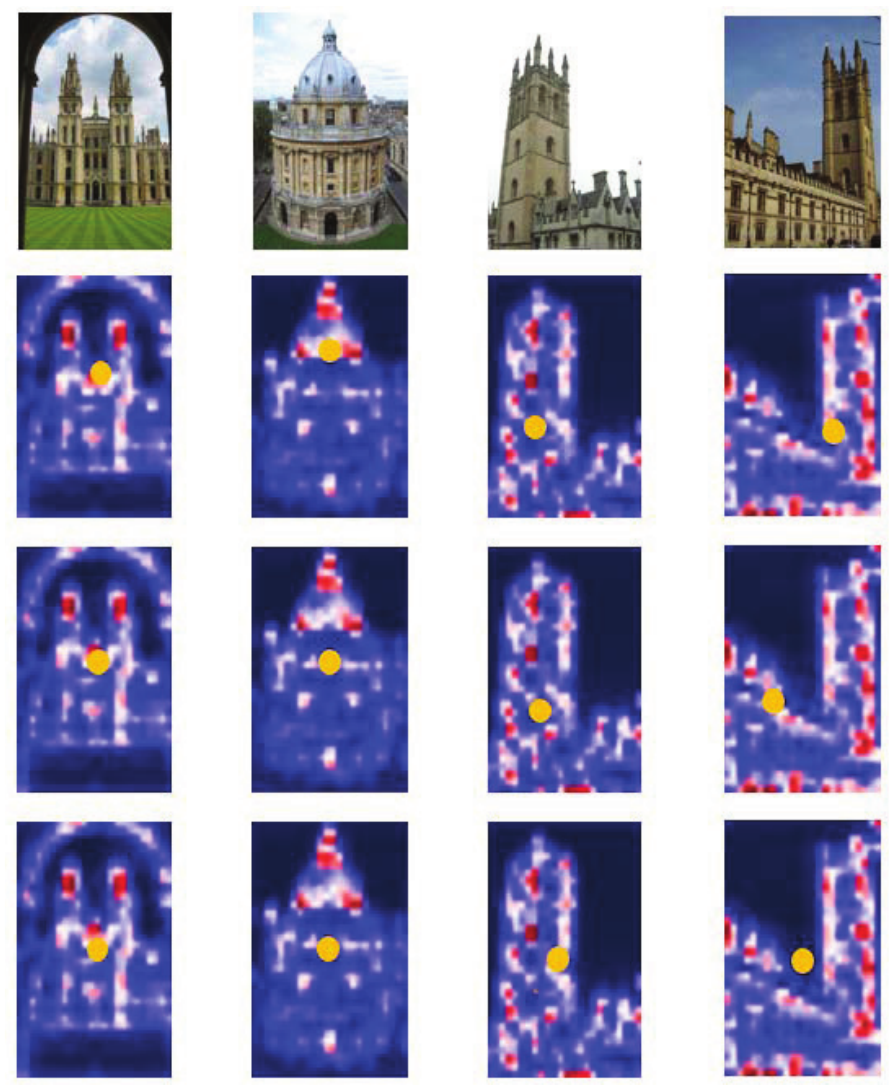,width=6.0cm}}
	\end{minipage}
	\caption{Original images and corresponding heat maps with Gaussian centers (yellow points) determined by selecting different values of $\alpha$. Centers in the 2nd, 3rd and 4th rows correspond $\alpha= 10\%$, $\alpha= 50\% $ and $\alpha= 100\% $, respectively.}
	\label{fig:cen}
\vspace{-0.1in}
\end{figure}

To determine  $({i_{0}},{j_{0}})$, we first introduce the matrix of aggregated response $S' \in {\mathcal{R}^{(W \times H)}}$ from all channels per spatial location:
\begin{eqnarray}
S^{'}_{(i,j)}{\rm{ = }}\sum\limits_{k = 1}^K {{X_{(i,j,k)}}} {\kern 1pt} {\kern 1pt} {\kern 1pt} {\kern 1pt} {\kern 1pt} \forall i=1,2...,W,{\kern 1pt} {\kern 1pt} {\kern 1pt} {\kern 1pt} j=1,2,...,H{\kern 1pt} {\kern 1pt} {\kern 1pt} {\kern 1pt} {\kern 1pt} {\kern 1pt} {\kern 1pt}
\end{eqnarray}
As an important finding, we notice $S^{'}$ contains the semantic information of the raw image. Fig.\ref{fig:cen} displays heat maps of $S'$ for some randomly selected images, where high and low values are denoted in red and blue, respectively. It clearly shows the large responses of $S^{'}$ correspond to salient regions. We sort out all regions and assign  geometrical center of top $\alpha$ large responses to be the center of Gaussian distribution. Therefore, the center $({i_{0}},{j_{0}})$ can be adaptively selected by $\alpha$.

Fig.\ref{fig:cen} shows geometrical centers obtained with three different $\alpha$ values. As displayed in Fig.\ref{fig:cen}, centers determined by the top 10\% large responses are more agreed with human vision perception. Besides, experiment results will further show that $\alpha = 10\%$ can give promising performance.

\begin{figure}[t]
	\begin{minipage}[b]{1.0\linewidth}
		\centering
		\centerline{\epsfig{figure=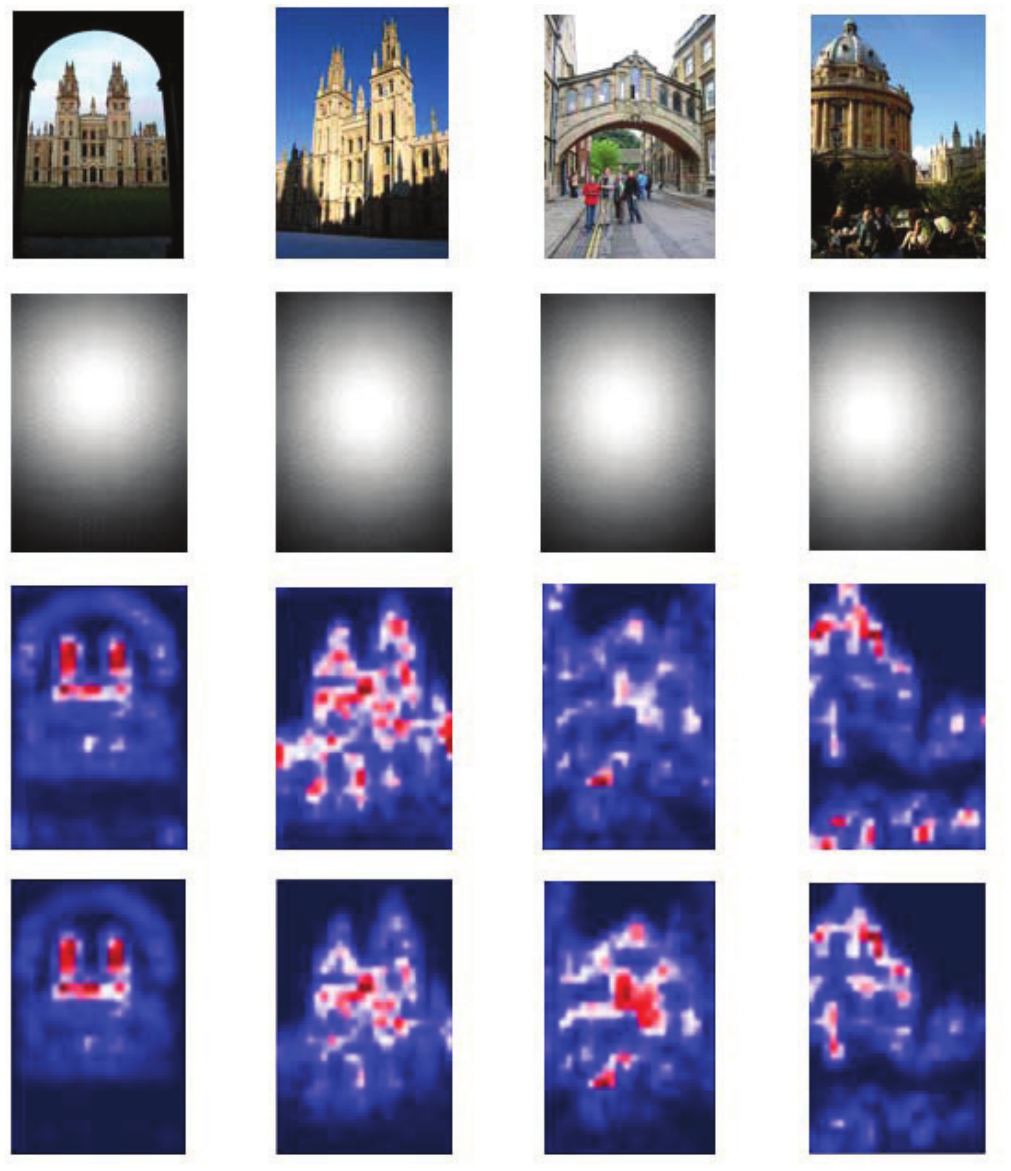,width=6.0cm}}
	\end{minipage}
	\caption{The responses of some examples activated by the adaptive Gaussian filter with $\alpha= 10\%$. The 1st row shows the original images, the 2nd row displays the Gaussian weight $S$, the 3rd row illustrates the heat maps $S^{'}$ and the 4th row demonstrates the responses of $S'\odot S$. }
	\label{fig:act}
\vspace{-0.1in}
\end{figure}

To examine the effect of spatial weighting, we visualize the active locations boosted by the aGaussian filter with four example images in Fig.\ref{fig:act}. As shown in Fig.\ref{fig:act}, RoI is strengthened while other regions are suppressed by our strategy.

\subsection{Element-value sensitive channel weighting}
BoW is a classical model for CBIR, where retrieval accuracy can be effectively boosted by inverse document frequency (idf) weighting. In the domain of deep feature based CBIR, different filters  usually activate different semantic content and generate corresponding feature maps. Inspired by the success of idf weighting, we want to differentiate channels in an image (Note that the standard idf weighting in the case of BoW is computed on a database)
and expect that channels with small aggregated values of feature maps are boosted. Accordingly, it is necessary to design channel weighting for deep convolutional features.

Kalantidis et.al \cite{kalantidis2016cross}  proposed a  channel weighting strategy based on the sparsity of feature maps. However, a feature map is a real-value but not a binary matrix. The element-value on each spatial location denotes the intensity of activation of the filter, therefore the sparsity does not make full use of all available information. Here, we propose a method to derive a new channel weighting based on  element-value. It is expected that similar images will have similar Gaussian filter responses for a given feature. For each channel, the item ${b_k}$ can be calculated as follows:
\begin{eqnarray}
{b_k} = {(\frac{{{\Omega _k}}}{{W \times H}})^2}{\kern 1pt} {\kern 1pt}
\end{eqnarray}
Obviously, the term ${b_k}$ is computed by summing the element-value of each feature map after Gaussian filtering.

\begin{figure}[t]
	\begin{minipage}[b]{.48\linewidth}
		\centering
		\centerline{\epsfig{figure=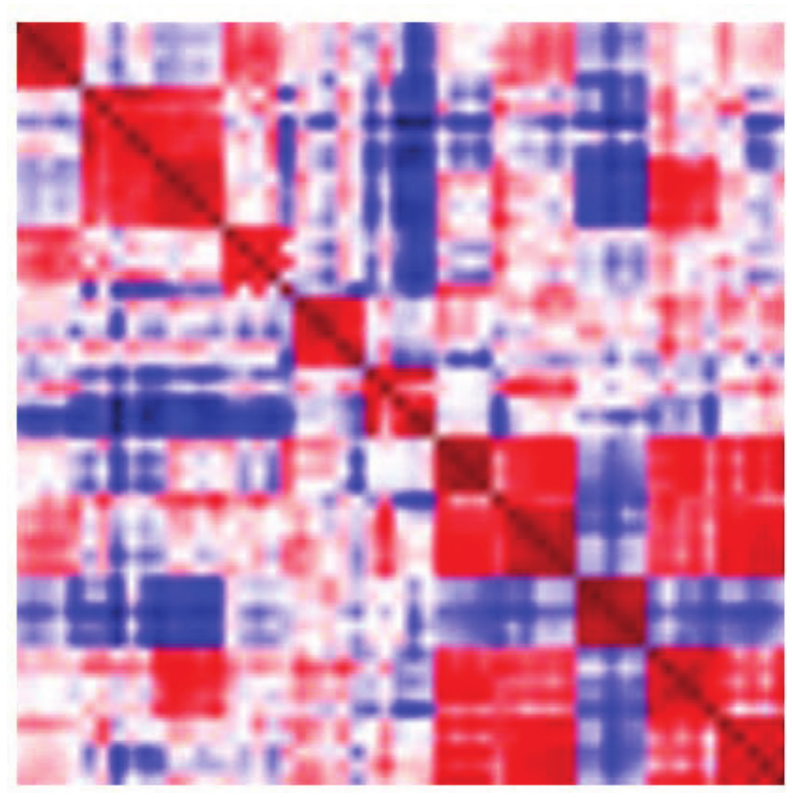,width=4.0cm}}
		\centerline{(a) Sparsity}\medskip
	\end{minipage}
	\begin{minipage}[b]{0.48\linewidth}
		\centering
		\centerline{\epsfig{figure=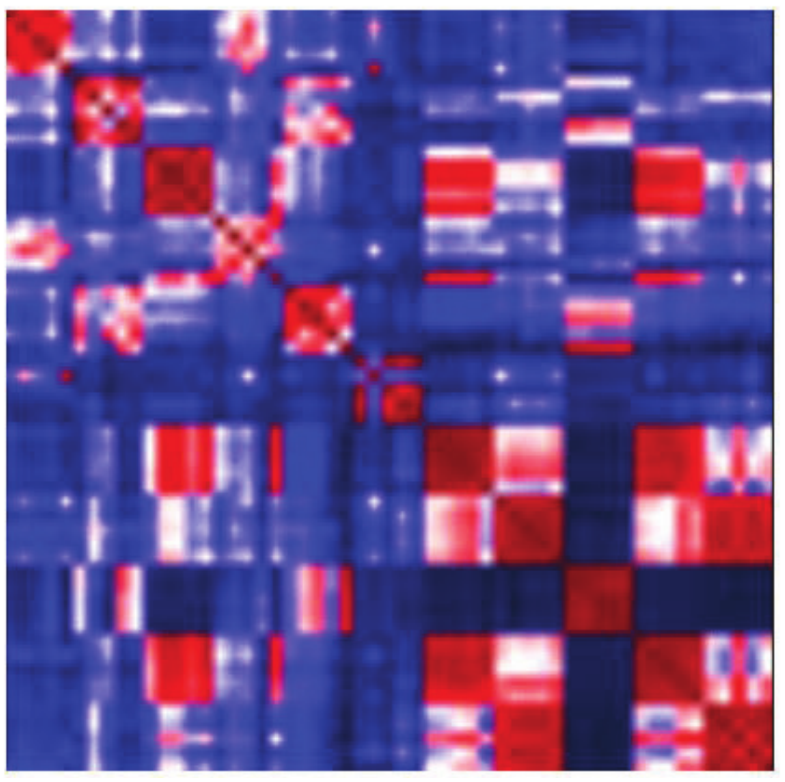,width=4.0cm}}
		\centerline{(b) Element-value}\medskip
	\end{minipage}
	\caption{The correlation of different vectors for the 55 images in the query-set of the Pari6K. }
	\label{fig:place}
\vspace{-0.1in}
\end{figure}

To compare with sparsity-sensitive weighting, we concatenate all ${b_k}$ into a $K$ dimensional vector $b = {[{b_1},{b_2},...,{b_K}]^T}$. Fig.\ref{fig:place} displays the pair-wise correlation of different vectors for all the images from the query set of the Paris6K dataset. This dataset contains 55 images, where each 5 images corresponds to a landmark of Paris and therefore images can be classified into 11 classes. As shown in Fig.\ref{fig:place}, both sparsity sensitive (Fig.3a) and our element-value sensitive vectors (Fig.3b) are highly correlated for images of a same landmark, but vectors computed by our strategy are less correlated than ones computed by the sparsity for images of different landmarks. Therefore, the information of the element-value sensitive vector is more discriminative than that of the sparsity sensitive vector. Besides, the feature with small average element-value could provide important information if, for example, only a small part of features have small average element-values for images in the same class. Hence, the element-value related term ${b_k}$ is used to replace the sparsity to compute channel weight as follows:
\begin{eqnarray}
{B_k} = \log (\frac{{K\varepsilon  + \sum\nolimits_c {b_c} }}{{\varepsilon  + b_k}}){\kern 1pt} {\kern 1pt} {\kern 1pt}
\end{eqnarray}
where a small constant $\varepsilon$ is added for numerical stability.

As the BoW model, deep convolutional features also suffer from the problem of visual burstiness \cite{jegou2009burstiness}. The introduction of channel weighting can alleviate this issue. Actually, channels with large average aggregated values correspond to CNN filters that give large response in many image regions. This implies there are some visual elements, which are spatial recurring and can negatively affect the retrieval accuracy. In our method,
small weights can be assigned to channels of such bursty CNN filters. Experiment results will illustrate its effectiveness for image retrieval.

\section{Experiments}\label{Experiments}
\subsection{Experimental Setting}
Our method is evaluated on five benchmark datasets, i.e., Oxford5K~\cite{philbin2007object} ($5,062$ building photos with $55$ queries including $11$ landmarks), Paris6K~\cite{philbin2008lost} ($6,392$ building photos with $55$ queries including $11$ landmarks), Oxford105K, Paris106K, and INRIA Holidays~\cite{jegou2010improving} ($1,491$ holiday photos with $500$ queries). Oxford105K and Paris106K are the extensions of Oxford5K~\cite{philbin2007object} and Paris6K~\cite{philbin2008lost} respectively, by adding other distracted $10$K images collected from Flickr. We employ the standard protocol consisting with other methods, i.e, using the cropped queries in Oxford5K(105K) and Paris6K(106K), adopting upright version of the images in Holidays. All deep features are extracted from the pool5 layer of the  VGG16~\cite{simonyan2014very} pre-trained in ImageNet, so the number of channels is $512$. The retrieval performance is measured by the mean Average Precision (mAP), which is defined as the average percentage of same class images in all retrieved images after evaluating all queries. Additionally, to fairly compare with other methods, we learn whitening parameters on Paris6K when testing on Oxford5K(105K), learn whitening parameters on Oxford5K when testing on Paris6K(106K) and INRIA Holidays.

\subsection{Impact of the parameter $\alpha$}
The proposed method has only one parameter that need to be evaluated, i.e., $\alpha$, determining the center of the spatial weighting matrix. In Section 2.2, we mentioned the large responses of $S'$ correspond to salient regions. However, "large" is a vague concept. Someone may think top $\alpha = 10\%$ responses in $S'$ are large, while others may think top 50\% responses in $S'$ are large. As different values of $\alpha$ would lead to different geometrical centers of the spatial weighting matrix, we empirically determine the value of $\alpha$ in practice.
\begin{table}[t]
	\begin{center}
		\caption{Performance of the spatial weighting with different values of $\alpha$ when tested  on Oxford5K. } \label{tab:alpha}
		\begin{tabular}{cc|cccccc}
			\hline
			\multicolumn{2}{c|}{$\alpha$}	&5\%	&10\% &15\%	&20\% &50\% &100\%\\
			\hline
			\multirow{3}{*}{Dim}
			&128 &63.0 &{\bfseries 63.3} &63.2 &63.2 &62.8 &61.2 \\
			&256 &67.9 &{\bfseries 68.4} &68.1 &67.7 &66.2 &63.4\\
			&512 &70.0 &70.4 &{\bfseries 70.6} &70.3 &69.2 &66.1\\
			\hline
		\end{tabular}
	\end{center}
\vspace{-0.1in}
\end{table}

Table~\ref{tab:alpha} shows the retrieval performance of Oxford5K~\cite{philbin2007object} in selecting different values of $\alpha$. In this experiment, we only apply aGaussian weighting without the eChannel weighting. According to Table~\ref{tab:alpha}, the excellent retrieval performance can be preserved when $\alpha$ is between 10\% and 20\%. While the retrieval performance begins to fall when $\alpha$ is larger than 50\%. Therefore, considering the trade-off between retrieval efficiency and accuracy, we empirically set $\alpha = 10\%$, which is also consistent with the results shown in Fig.\ref{fig:cen}.

\subsection{Impact of different weighting schemes}

\begin{figure}[t]
	\begin{minipage}[b]{1.0\linewidth}
		    \centering
			\centerline{\epsfig{figure=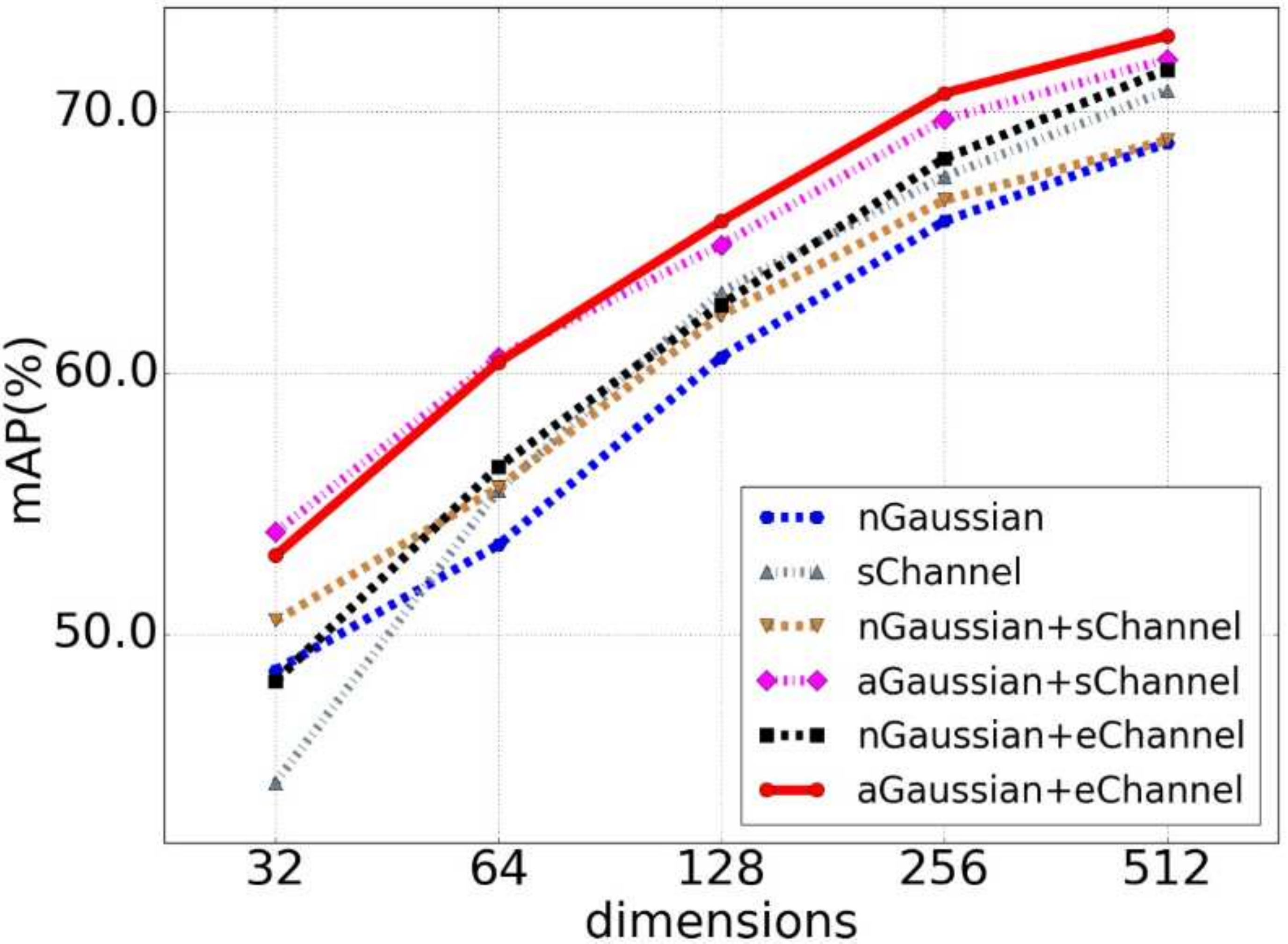,width=6.3cm}}
	\end{minipage}
	\caption{Comparison of retrieval accuracy for different weighting combinations tested on Oxford5K.}
	\label{fig:mAP}
\vspace{-0.1in}
\end{figure}

In our method, the aGuassian and eChannel weighting strategies are proposed to co-weight the deep convolutional features, and their effectiveness should be verified by the experiment. For spatial weighting, we compare the aGuassian weighting with the normal Gaussian (nGaussian) weigthing, which corresponds the aGaussian weighting with $\alpha=100\%$. For channel weighting, we compare the eChannel weigthing with the sparsity channel (sChannel) weighting. As the spatial and channel weight are independently, they can be individually or simultaneously applied to deep convolutional features. Six groups of weighting combination are tested on Oxford5K under different $K'$ for image retrieval. Fig.\ref{fig:mAP} displays the retrieval accuracy for these groups of combination. As we can see, both aGaussian and eChannel weighting can improve the accuracy of image retrieval. What's more, the combination of them can further contribute to the improvement of retrieval accuracy. Therefore, both the proposed weighting are validated for the aggregation of convolutional features.

\begin{figure*}[t]
	\begin{minipage}[b]{1.0\linewidth}
		\centering
		\centerline{\epsfig{figure=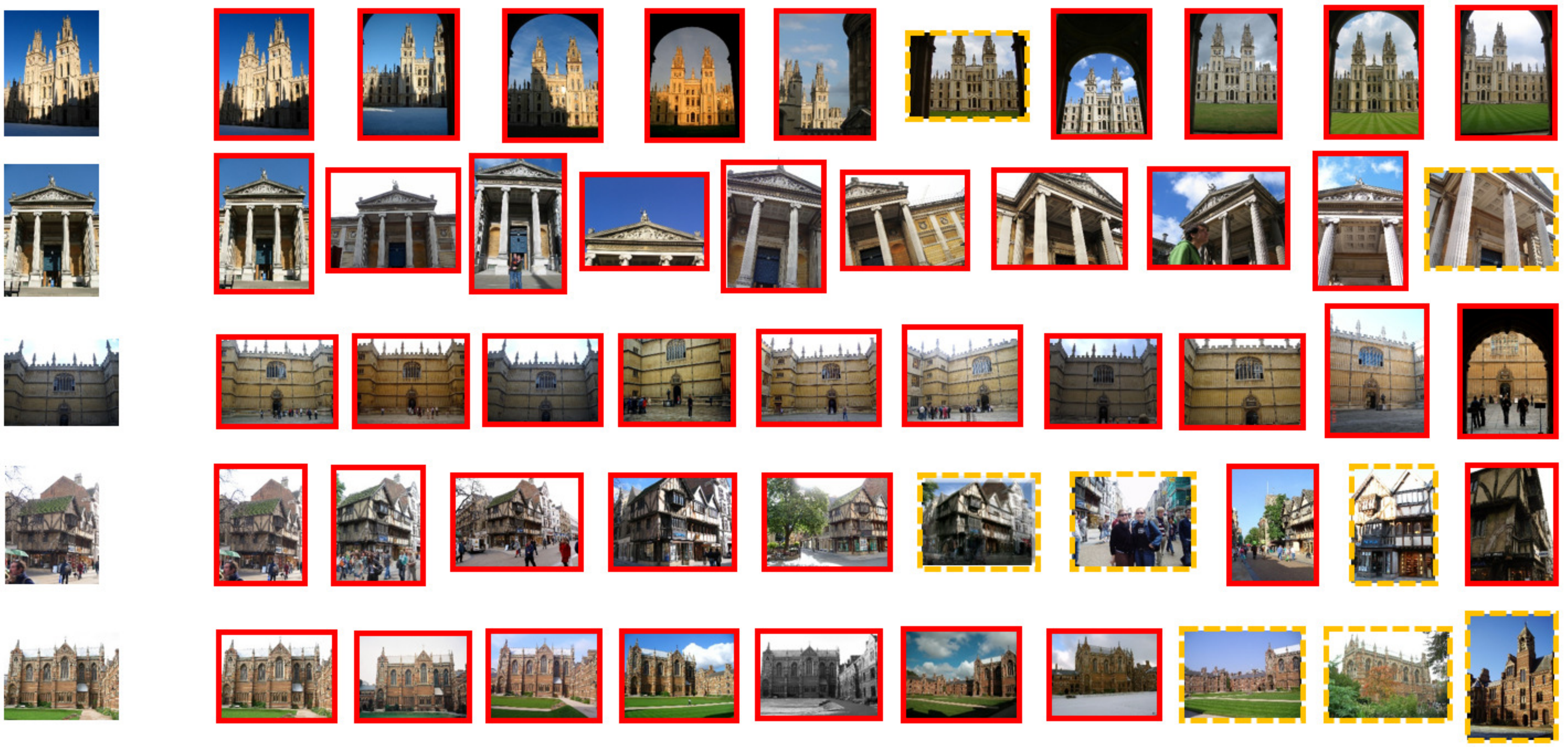,width=16cm}}
	\end{minipage}
	\caption{Top 10 results for randomly selected five queries of Oxford5K, using a 512-dimensional global vector to represent each image. The query images are displayed on the leftmost place. The true and false results are marked with red and yellow dotted borders respectively.}
	\label{fig:example_results}
\vspace{-0.1in}
\end{figure*}
\subsection{Comparison with the-state-of-the-art}

Table~\ref{tab:cap} compares the retrieval performance of the proposed method with several relevant methods without fine-tuning, namely CroW~\cite{kalantidis2016cross}, Neural code~\cite{babenko2014neural}, R-MAC~\cite{tolias2016particular}, SPoC~\cite{babenko2015aggregating} and Razavian et al~\cite{sharif2014cnn}.
As shown, we compare them on five benchmark datasets with different dimensions. According to the table, our method can achieve overall the best performance among the compared methods. Particularly, in the datasets of Oxford5K(105K) and Holidays, the proposed method is at least $2\%$ accurate compared with the other methods using $512$ dimensional features. Fig.~\ref{fig:example_results} illustrates several randomly selected retrieval results of the Paris6K.

\begin{table*}[t]
	\begin{center}
		\caption{Performance comparison (in mAP) with recent deep feature based image retrieval methods. Our method consistently outperforms all of them.} \label{tab:cap}
		\begin{tabular}{ccccccc}
			\hline
			Method	&Dim	&Paris6K	&Oxford5K	&Paris106K	&Oxford105K	&Holidays\\
			\hline
			NetVLAD~\cite{arandjelovic2016netvlad} &512 &74.9 &67.6 &-- &-- &86.1\\
			CroW~\cite{kalantidis2016cross}	&512	&79.7	&70.8	&72.2	&65.3	&85.1 \\
			Neural code~\cite{babenko2014neural}&	512	&--	&55.7	&--	&52.2	&78.9\\
			R-MAC~\cite{tolias2016particular}	&512	&{\bfseries 83.0}	&66.9	&75.7	&61.6	&--\\
			Razavian~\cite{sharif2014cnn}	&512	&67.4	&46.2	&--	&--	&74.6\\
			Our method	&512	&{\bfseries 83.0}	&{\bfseries 72.8}	&{\bfseries 76.3}	&{\bfseries 68.1}	&{\bfseries 87.4}\\
			\hline
			NetVLAD~\cite{arandjelovic2016netvlad} &256 &73.5 &63.5 &-- &-- &84.3\\
			SPoC~\cite{babenko2015aggregating}	&256	&--	&53.1	&--	&50.1	&80.2\\
			R-MAC~\cite{tolias2016particular}	&256	&72.9	&56.1	&60.1	&47.0	&--\\
			Neural code~\cite{babenko2014neural}	&256	&--	&55.7	&--	&52.4	&78.9\\
			CroW~\cite{kalantidis2016cross}	&256	&76.5	&68.4	&69.1	&63.7	&85.1\\
			Our method	&256	&{\bfseries 80.5}	&{\bfseries 70.7}	&{\bfseries 74.0}	&{\bfseries 66.5}	&{\bfseries 87.2}\\
			\hline
			NetVLAD~\cite{arandjelovic2016netvlad} &128 &69.5 &61.4 &-- &-- &82.6\\
			CroW~\cite{kalantidis2016cross}	&128	&74.6	&64.1	&67.0	&59.0	&82.8\\
			Neural code~\cite{babenko2014neural}	&128	&--	&55.7	&--	&52.3	&78.9\\
			Our method	&128	&{\bfseries 77.9}	&{\bfseries 65.8}	&{\bfseries 70.5}	&{\bfseries 61.4}	&{\bfseries 85.7}\\
			\hline
		\end{tabular}
	\end{center}
\end{table*}

We also compare our method with a fine-tuned method NetVLAD~\cite{arandjelovic2016netvlad}.
According to Table~\ref{tab:cap}, one can find that our method outperforms NetVLAD on all the test datasets with diverse dimensions by a significant margin.
When comparing with the best reported results of other very recent two fine-tuning works~\cite{radenovic2016cnn, gordo2016deep},
our results seem not optimal.
However, these fine-tuning methods~\cite{arandjelovic2016netvlad, radenovic2016cnn, gordo2016deep}  heavily rely on manual annotation and training correction. Therefore, advantages of our method can be further highlighted for the problems where the fine-tuning methods cannot work well.

\section{Conclusion}\label{Conclusion}
In this paper, we propose an unsupervised aggregation method for image retrieval using convolutional features. The key characteristics is that we design an adaptive Gaussian filter and an element-value sensitive channel vector to co-weight deep convolutional features extracted from the pre-trained CNN. The former can highlight the retrieval objects of the image and the latter can ease the burstiness phenomenon in deep local features. We analyze the motivation of these weighting schemes and prove their effectiveness by experiments.

Experiments on five benchmark datasets demonstrate our method outperforms other very recent aggregation methods based on off-the-shelf deep features without needing a long dimension vector to represent each image. It is worth noting our unsupervised aggregation method is quite suitable and effective under the situation where the related training dataset is difficult to collect and the retrieval dataset is quite large which requires a lot of storage space.

\section*{\normalsize{Acknowledgements}}
This work was supported by National Natural Science Foundation of China (NSFC) (Grant Nos.~61603289 and 61573273);
China Postdoctoral Science Foundation (Grant No.~2016M602823); Postdoctoral Science Foundation of Shaanxi;
and Fundamental Research Funds for the Central Universities (Grant No.~xjj2017118).
\vspace{-0.1in}

\bibliographystyle{IEEEbib}
\bibliography{camera-ready_icme2018template}

\end{document}